\documentclass[runningheads]{llncs}
\usepackage{graphicx}
\usepackage{tikz}
\usepackage{comment} 
\usepackage{amsmath,amssymb} % define this before the line numbering.
\usepackage{color}
\usepackage[lofdepth,lotdepth]{subfig}
\usepackage{subfig}
\usepackage{enumitem}
\usepackage{xcolor}
\usepackage{tabularx}
\usepackage{booktabs}
\usepackage{arydshln}
\usepackage[numbers,sort]{natbib}

\usepackage{subfig}
\setlist[enumerate]{itemsep=0mm}

\begin{document}
% \renewcommand\thelinenumber{\color[rgb]{0.2,0.5,0.8}\normalfont\sffamily\scriptsize\arabic{linenumber}\color[rgb]{0,0,0}}
% \renewcommand\makeLineNumber {\hss\thelinenumber\ \hspace{6mm} \rlap{\hskip\textwidth\ \hspace{6.5mm}\thelinenumber}}
% \linenumbers
\pagestyle{headings}
\mainmatter
\def\ECCVSubNumber{5685}  % Insert your submission number here

\title{Directional Temporal Modeling for Action Recognition} % Replace with your title

% INITIAL SUBMISSION 
\begin{comment}
\titlerunning{ECCV-20 submission ID \ECCVSubNumber} 
\authorrunning{ECCV-20 submission ID \ECCVSubNumber} 
\author{Anonymous ECCV submission}
\institute{Paper ID \ECCVSubNumber}
\end{comment}
%******************

% CAMERA READY SUBMISSION
%\begin{comment}
\titlerunning{Directional Temporal Modeling for Action Recognition}
% If the paper title is too long for the running head, you can set
% an abbreviated paper title here
%
% \author{Xinyu Li\orcidID{0000-1111-2222-3333} \and
% Bing Shuai \orcidID{1111-2222-3333-4444} \and
% Joe Tighe \orcidID{2222--3333-4444-5555}}

\author{Xinyu Li \and
Bing Shuai \and
Joseph Tighe}

\authorrunning{X. Li et al.}
% First names are abbreviated in the running head.
% If there are more than two authors, 'et al.' is used.
%
\institute{
Amazon Web Service \\
\email{\{xxnl,bshuai,tighej\}@amazon.com}}
%\end{comment}
%******************
\maketitle

\begin{abstract}
Many current activity recognition models use 3D convolutional neural networks (e.g. I3D, I3D-NL) to generate local spatial-temporal features. However, such features do not encode clip-level ordered temporal information.
In this paper, we introduce a channel independent directional convolution (CIDC) operation, which learns to model the temporal evolution among local features.
By applying multiple CIDC units we construct a light-weight network that models the clip-level temporal evolution across multiple spatial scales. Our CIDC network can be attached to any activity recognition backbone network.   
We evaluate our method on four popular activity recognition datasets and consistently improve upon state-of-the-art techniques. 
We further visualize the activation map of our CIDC network and show that it is able to focus on more meaningful, action related parts of the frame.
\end{abstract}

%%%%%%%%% BODY TEXT
\section{Introduction}

Action recognition has made significant progress in recent years \cite{lin2018temporal,martinez2019action,karpathy2014large,feichtenhofer2018slowfast,wang2018non}, with most of these methods leveraging 3D convolutions to learn spatial-temporal features.
Most operate by taking as the input a video clip (a set of contiguous frames extracted from a video), and passing it through a 3D feature backbone. After the final convolutional block, the temporal dimension of feature map is typically down-sampled by a factor of $t$ (e.g. 8). At this point the feature map at each temporal position represents a $t$ frame sub-clip. The spatial-temporal feature map is passed to a global average pooling layer to summarize its salient features, and this pooled feature is used to derive the activity class label.

Even though 3D feature extraction backbones \cite{wang2018non}) and their derivatives (e.g. I3D-NL \cite{wang2018non}) have proven to be effective, there are two major issues: (1) while the 3D convolutions in these networks have a temporal receptive field that spans the full clip, the effective receptive fields have been shown to actually be quite local \cite{luo2016understanding} and thus lack full clip level motion understanding and (2) the temporal ordering/relationships between sub-clips is lost by the global average pooling of the feature maps.

\begin{figure}[t]
	\begin{center}
		\includegraphics[width=\textwidth]{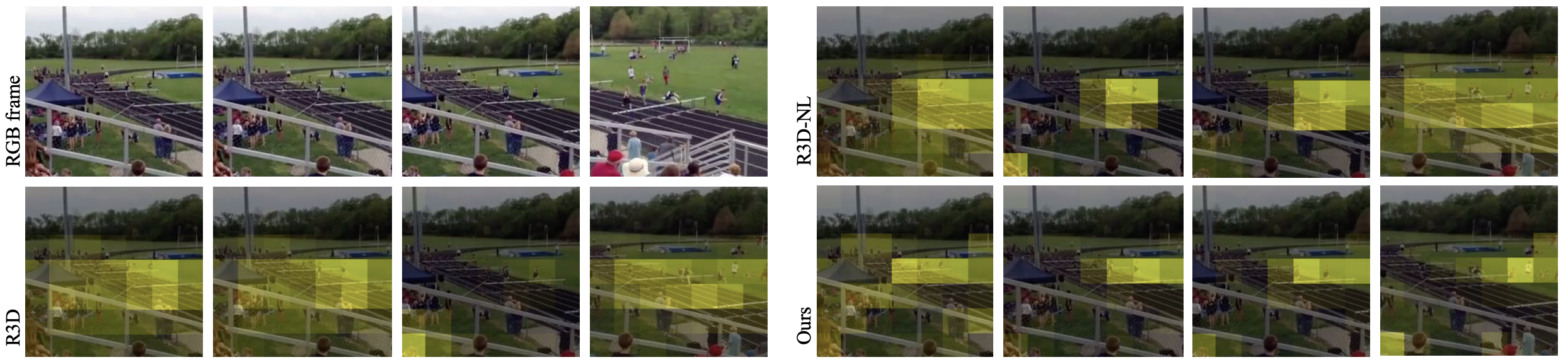}
	\end{center}
	\caption{
	We show a video clip (32 frames) and the spatial activation maps for the representative frame of every sub-clip (8 frames). I3D and I3D-NL not only activates related image regions of ``\texttt{althelets hurdling}", but also activates ``\texttt{audiences moving}" in the background. In contrast, our proposed model has clip-level motion understanding, thus it largely focuses on image region of interest that explains the action ``\texttt{althelets hurdling}". Video examples are from Kinetics-400 dataset.
	}
	\label{fig:1}
\end{figure}

To this end, we propose a novel channel independent directional convolution (CIDC) operation that captures temporal order information between sub-clips. Specifically, our CIDC unit encodes a feature vector that progressively aggregates the backbone features over the full extent of the input video clip. In other words, the first element in our method's output represents only the first few frames of the input clip, the middle element: the first half of the clip, and the last element: the full clip. We perform this operation bidirectionally (both forwards and backwards) to better capture the complete temporal order information of the video clip.

We use our CIDC unit to construct a light-weight CIDC network that can be attached to any activity recognition backbone (e.g. I3D, I3D-NL) to explicitly aggregate clip-level motion information. Our CIDC network is formed by stacking multiple CIDC units on top of the last three blocks of any action recognition backbone network. Our network is able to aggregate multiple spatial-temporal feature maps from different scales to effectively encode different types of motion.

As can be seen in Figure \ref{fig:1}, although the I3D and I3D-NL feature encoders are able to focus on the image areas correlated with the activity, they also activate on image regions where irrelevant object motion happens (i.e audience moving in the background). In contrast, the proposed network is able to understand the longer-term clip-level motion, therefore it precisely localizes the key area of the video for the action of interest (i.e. athletes hurdling).

% \joe{need to have nice crisp summary of contributions}
We test our multi-scale CIDC network on four datasets: UCF-101 \cite{soomro2012dataset}, HMDB-51 \cite{kuehne2011hmdb}, Something-Something V2 \cite{goyal2017something} and Kinetics-400 \cite{kay2017kinetics}. 
Our model consistently improves state-of-the-art activity recognition models, demonstrating the effectiveness of the proposed method. 
% We also perform careful ablation studies to validate the benefits of each components of the proposed model. Moreover, we visualize the spatial activation maps of different action recognition networks, which confirm our intuition that proposed model differs from state-of-the-art networks to model temporal associations. 
%The results show that by attaching our CIDC network to R2D backbone, we were able to achieve similar performance compared with I3D backbone trained on same amount of data, showing that the global spatial-temporal feature helps wth recognition. 
%The results show that our CIDC network attached to I3D backbone outperforms other spatiao-temporal modeling methods including no-local \cite{wang2018non} and TSM \cite{lin2018temporal} on multiple datasets. \joe{repetitive}
%We finally visualized the feature learned by different networks and discussed the difference between different activity recognition strategy based on visualization. 
Overall, our contributions are:
\begin{enumerate}[nosep]
	\item A novel {channel independent directional convolution} (CIDC) unit that aggregates features temporally and maintains the relative temporal order in the generated feature. 
	\item A multi-scale CIDC network that learns video clip-level temporal feature from different spatial and temporal scales. 
	\item An in-depth analysis and visualization of CIDC network that shows it is able to leverage clip-level temporal association to better focus on action-related features.
\end{enumerate}

The rest of this paper is organized as follows. 
Section \ref{sec:related_work} discusses the related work. 
Section \ref{sec:meth} elaborates the technical details of the proposed multi-scale CIDC networks. 
Section \ref{sec:exp} presents our experimental results. 
Section \ref{sec:discussion} concludes the paper.

%------------------------------------------------------------------------
\section{Related Work}
\label{sec:related_work}
\paragraph{Feature representation for activity recognition}

In order to represent an action, the video-level feature encoder needs to summarize the information about the objects, scenes as well as target object motions in the videos. First, researchers use ConvNet-2D \cite{karpathy2014large,simonyan2014two,wang2016temporal,feichtenhofer2016convolutional,lin2018temporal} to extract feature for every frame, and aggregate frame-wise features to video-level feature.  In order to encode motion information, two stream ConvNet-2D \cite{feichtenhofer2016convolutional,wang2016temporal, karpathy2014large, simonyan2014two} are used, in which optical flow images \cite{perez2013tv} are directly taken as input to complement visual appearance information from RGB-stream Convnet-2D. Recently, ConvNet-3D \cite{carreira2017quo,taylor2010convolutional,tran2018closer,xie1712rethinking, tran2015learning,wang2018non,feichtenhofer2018slowfast,martinez2019action} extends ConvNet-2D to spatial-temporal domain, handling both spatial and temporal dimensions similarly. It achieves successes on action recognition in terms of its model efficiency and model capacity. TSM \cite{lin2018temporal} and TAM \cite{fan2019blvnet} propose to perform temporal modeling by shifting the feature channel along temporal dimension. 
SlowFast \cite{feichtenhofer2018slowfast} models the action with a motion (fast) branch and a visual appearance (slow) branch, and it achieves the state-of-the-art performance.
LFB \cite{wu2019long} adopts a self-attention block to aggregate video-level features \cite{wu2019long}.    
In this paper, we propose a novel light-weight CIDC network to learn clip-level temporal association, which is important for action recognition. More importantly, the proposed CIDC network is complementary to previous works that usually captures local spatio-temporal information.

\paragraph{Temporal modelling in activity recognition.}
Temporal modelling is considered to be essential for action recognition \cite{lin2018temporal, feichtenhofer2018slowfast,wu2019long}. LSTMs \cite{hochreiter1997long} are firstly used to model temporal associations among features extracted by 2D networks \cite{li2017progress,yue2015beyond,donahue2015long,li2018videolstm}. However, its results are not as good as expected in helping activity recognition \cite{lin2018temporal, feichtenhofer2018slowfast,wu2019long}. 
The  temporal  rank  pooling  is  another  way  to  model the evolution of actions in sequential order \cite{fernando2016rank,bilen2016dynamic,fernando2015modeling}. Temporal rank pooling requires flattened features, which may compromises the spatial arrangement of the features, which makes it not feasible to insert the temporal rank pooling in the middle of the network.
Furthermore, self-attention block \cite{vaswani2017attention,shen2018disan} is used to aggregate spatial-temporal features, and it helps improving the action recognition performance \cite{wang2018non,girdhar2019video}. In this paper, we propose channel independent directional convolution (CIDC) network to model clip-level temporal associations. We  empirically show that our CIDC network outperforms LSTM and self-attention block in improving action recognition accuracy. 

\section{Methodology}
\label{sec:meth}

A typical action recognition network takes as input a video clip (a set of $n$ contiguous video frames), and passes it through a 3D feature backbone (e.g. I3D, I3D-NL \cite{wang2018non}) to create a feature map ($\mathbf{F}^{C \times T \times W \times H}$), where $T$ indicates the temporal length, $C$ the number of channels and $W \times H$ the width and height.  
%We denote the output feature map of the backbone network as $\mathbf{F}^{C \times T \times W \times H}$, in which  $T$ indicates the temporal length, $C$ is the number of channels and  $W \times H$ the width and height. 
While the 3D convolutions that create this feature map do have a temporal receptive field that spans the full clip ($T$), the effective receptive fields of convolutional features have been shown to generate local, sub-clip descriptors \cite{luo2016understanding}, rather descriptors that capture the long term temporal evolution of the full clip. Unfortunately, most networks at this point perform global average pooling over the spatial-temporal dimensions, thus throwing away any information of how the action evolves over the entirety of the clip. In this section we present our method to overcome this limitation by explicitly modeling the temporal evolution over the full clip. 

%Thus the resulting feature after the average pooling that follows the 3D convolutional feature extractor is not able to capture long term temporal relationships.In this case, the feature map is temporally down-sampled by a factor of $\frac{n}{T}$ (e.g. 8). We note that the $t^{\text{th}}$ temporal slice of the feature map $\mathbf{F}_t \in \mathbb{R}^{C \times W \times H}$ summarize the information of the $t^{\text{th}}$ sub-clip ($\frac{n}{T}$ contiguous frames within the video clip), but the global average pooling that is performed in most action recognition systems throws away the inter-sub-clip relationships. In this section we present our method that preserves this temporal information.
\subsection{Channel Independent Directional Convolution}
\label{sec:CIDC}

To explicitly encode the temporal evolution of the clip, we introduce our novel {\bf directional convolution} operation. This operation can be thought of as set of 2D convolution over the spatial dimensions that progressively add more temporal context. The first convolution of this set only operates on the first temporal element of the clip volume. The subsequent spatial convolutions progressively incorporate more of the temporal extent of the clip until the final convolution considers the whole clip. Considering that different channels in the feature map represent different visual components, applying the same directional convolution across all channels restricts the temporal modelling capacity of the network. Thus, we instead apply this directional convolution independently per channel, and we refer to the complete operation as a {\bf channel independent directional convolution} or CIDC. In the following section we present how we implement our proposed CIDC.

\subsubsection{Channel Independent Directional Convolution Implementation}
\label{sec:directed_convolution}
We implement our proposed CIDC operator as 2D convolutions that operate on each channel ($C$) independently and treat the temporal dimension ($T$) as the channels normally would be. 
Consider the input feature map $\mathbf{F}_c$ as the feature map for the $c^\text{th}$ channel where $\mathbf{F}_c \in \mathbb{R}^{T \times W \times H}$. We convolve it with $T'$ filters $\mathbf{w}^c =[\mathbf{w}_1^c; \ldots \mathbf{w}_t^c; \ldots; \mathbf{w}_{T'}^c]$ to produce a feature map $\mathbf{F}_c'\in \mathbb{R}^{T'\times  W\times H}$:

% we perform the following operations:
% This operation is denoted as

%\begin{equation}
%	\mathbf{F}'=  \text{concat} \left{(} \mathbf{w}_t(1:t) * \mathbf{F}(1:t)\right{)}_{t\in{1..T}}
%\label{eq:temp_conv}
%\end{equation}

\begin{equation}
	\mathbf{F}_c'= \text{concat} \left.{)} \mathbf{w}_t^c * \mathbf{F}_c \right.{)}_{t\in{1..T}} 
\label{eq:temp_ind_conv}
\end{equation}

\noindent and then concatenate all $c$ features maps to produce $\mathbf{F}'\in \mathbb{R}^{T'\times C\times W\times H}$, which is our generated spatial-temporal feature map.
%\noindent where $*$ denotes the convolution operation, $\text{concat}()$ represents the tensor stacking operation and $\mathbf{w} =[\mathbf{w}_1; \ldots \mathbf{w}_t; \ldots; \mathbf{w}_{T'}]$ is the set of $T'$ $1 \times 1$ convolution kernels ($\mathbf{w}_t \in \mathbb{R}^{T \times 1 \times 1 \times 1}$). Here the same weights $\mathbf{w}$ are shared across all $c$ channels of feature map $\mathbf{F}$.
%
%$\mathbf{F}'\in \mathbb{R}^{T'\times C\times W\times H}$ is the generated spatial-temporal feature map, which is rearranged to $\mathbf{F'} \in \mathbb{R}^{C\times T' \times W \times H}$.
This brings us close to the operator outline in section \ref{sec:CIDC}, but if we apply such a convolution naively over feature maps we do not have any guarantees that it will capture the temporal evolution of the full clip.

To create our CIDC operator we force the upper triangle of each $\mathbf{w}^c$ to be zeros. By doing so the output features gradually represent larger portions of the video: the first element only has the context of the first sub-clip of the video clip, the center element has the context of the first half of the video clip and the last element has the context of the whole video clip. The operation is straight forward to implement when $T=T'$ but when this condition is not met (when we perform temporal re-sampling) it is less obvious. We cover this case in detail in section \ref{sec:impl_dets}, as it turns out the efficient implementation of this operation can deal with this dimension miss match trivially.

Our method has the added benefit that by changing the output temporal dimension $T'$, we are able to``softly'' manipulate the temporal dimension.
This is important, as it avoids the significant information loss due to temporal pooling or temporal convolution with strides that current 3D backbones use.
%But such methods assume the useful information is uniformly distributed across all frames, which is not always true. Our methods can avoid the information loss caused by pooling. 

\begin{figure}[t]
	\begin{center}
		\includegraphics[width=1\linewidth]{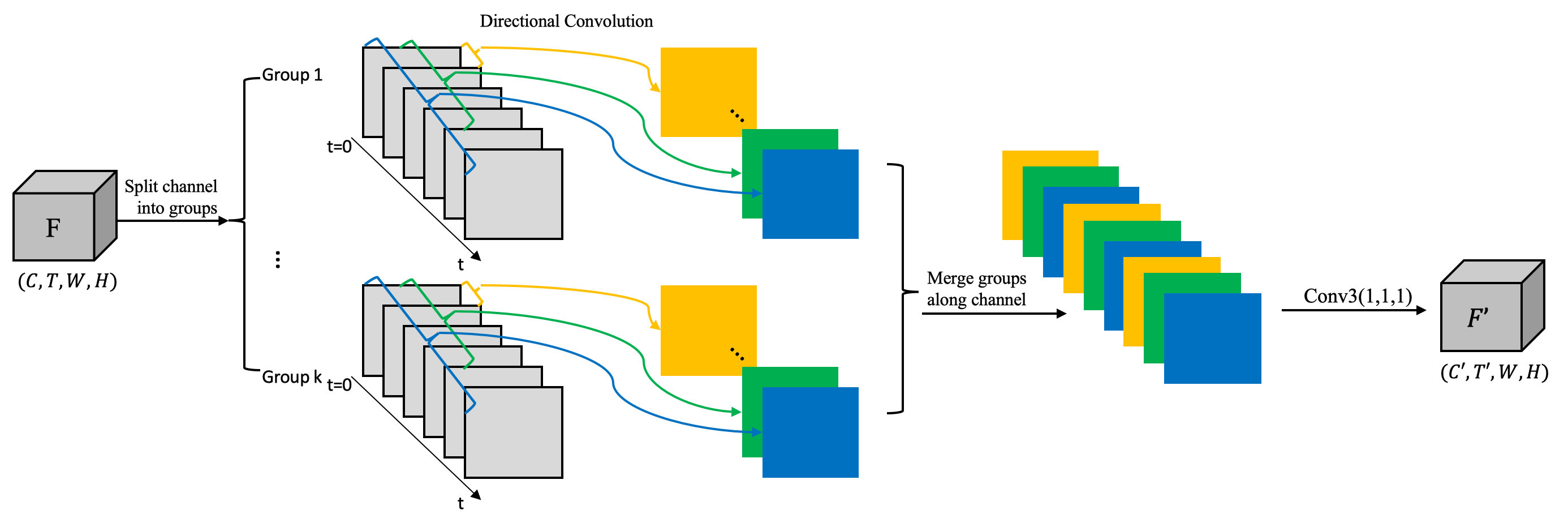}
	\end{center}
	\caption{Graphical illustration of a single Channel Independent Directional Convolution (CIDC) unit. 
	%\bing{Why the order of tensor is different from the main text.}
	}
	\label{fig:2}
\end{figure}

To have an efficient implementation of the above algorithm, instead of splitting the operation into $C$ separate operators, we combine the temporal and channel dimensions and maintain the channel independence by using grouped convolutions \cite{krizhevsky2012imagenet,xie2017aggregated}  where the number of groups is $C$. In this work we only use $1\times 1$ filter sizes for $w$ as our main focus is on the temporal, not spatial modeling but there is nothing that specifically restricts one to this choice. After this operation, we also apply a standard channel-wise 3D convolution with kernel size of 1 to learn the semantic features after temporal aggregation (Figure \ref{fig:2}). 
%This allows the network to learn on the important features using clip-level temporal information.

\subsection{Multi-scale CIDC Network}
\label{sec:CIDC_net}

Although the spatial-temporal feature map $\mathbf{F}$ encodes sub-clip level motion information, it lacks longer-time clip-level motion understanding. To encode this, it is important to get rid of distracting motion information from irrelevant objects. Even though in some case, background objects or their motion information can help action recognition, we argue that the model should focus on the motion of the target objects in order to achieve deeper video understanding (e.g. spatial-temporal action detection, etc.).

To this end, we construct our CIDC network (Figure \ref{fig:3}) by stacking multiple CIDC units and attaching it to a backbone network. The resulting output feature map of our CIDC network is compact, has temporally ordered information and more importantly, it aggregates the clip-level temporal information. Therefore, the network can leverage such information to activate image regions that can consistently explain the actions (athletes \texttt{hurdling} in Figure \ref{fig:1}) in the video clips rather than to attend to short-term background motions (audiences \texttt{moving} due to camera motion in Figure \ref{fig:1}).
The output of our network ($\mathbf{F}_{CIDC}$) has the same dimensions ($C \times T \times W \times H$) as the backbone output $\mathbf{F}$ before average pooling. 
 
We expect these two features to be complementary, and we fuse $\mathbf{F}$ and $\mathbf{F}_{CIDC}$ before they are passed to a final  classification layer. 
We define,
\begin{equation}
    \mathbf{F}_{out} = \mathcal{S}(\mathbf{F}, \mathbf{F}_{CIDC})
\label{eqn:fusion}
\end{equation}
\noindent where $\mathcal{S}$ is the fusion function.  We explore the following fusion functions: (1), concatenation $\mathbf{F}_{out} = [\mathbf{F}; \mathbf{F}_{CIDC}]_d$, where $d$ indicates the dimension along which the concatenation operation happens; (2), summation $\mathbf{F}_{out} = \mathbf{F} + \mathbf{F}_{CIDC}$.

\subsubsection{Multi-scale aggregation}
\begin{figure}[t]
	\begin{center}
    	\includegraphics[width=0.95\linewidth]{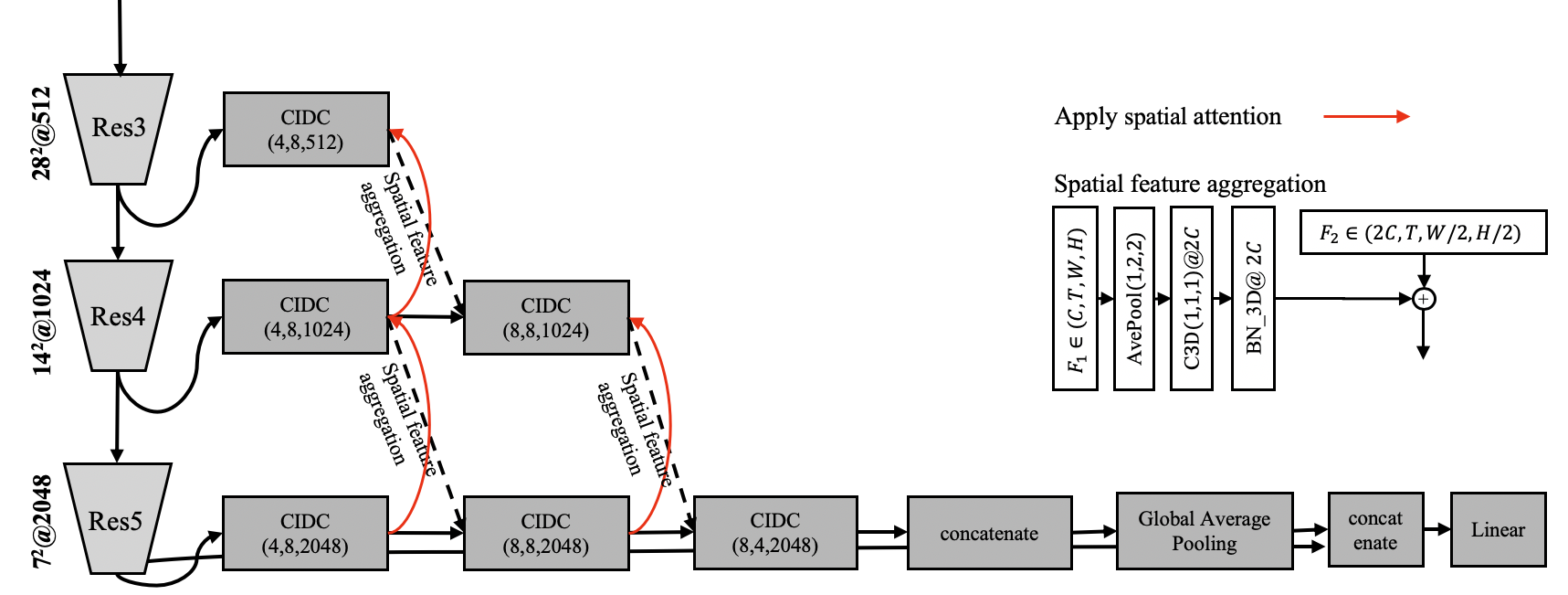}
	\end{center}
	\caption{
% 	\joe{update figure with new notation: CIDC($T_1, T_2, C$). Also I think you should rotate the ResX feature extractors to point down and draw lines branching no coming from the side of the block}
	    The architecture of multi-scale CIDC network, in which it aggregates multiple spatial-scale feature maps. 
	    The dotted line arrows denotes spatial feature aggregation operation from higher-resolution feature maps to lower-resolution ones.
	    The solid line arrow indicates the spatial attention propagation operation.
	    CIDC($T_1, T_2, C$) refers to a CIDC unit with $T_1$ input temporal length, $T_2$ output temporal lenght and $C$ channels.}
% 		The CIDC network integrate the feature from from large spatial-temporal scale to small spatial-temporal sclase. 
% 		The dot line arrow shows merging feature from high spatial resolution to low-spatial resolution at same temporal scale.
% 		The solid line arrow shows temporal feature aggregation using CIDC.
% 		The dash line arrow denotes apply the spatial attention at same temporal scale.
% 		CIDC($T1,T2@C$) denotes the CIDC unit with $T1$ input temporal dimension, $T2$ output temporal dimension and with $C$ channels.
	\label{fig:3}
\end{figure}

Inspired by HRNet \cite{sun2019high,sun2019deep}, our CIDC network consists of multiple CIDC branches attached to different scale feature maps, with cross-scale links.
Instead of applying bi-directional cross scale aggregation, as in HR-Net, we only pass early stage feature maps (with higher resolution) to later stage feature maps (with lower resolution).  
To achieve this aggregation, our feature aggregation unit performs the necessary dimensionality reduction to ensure that feature vectors are compatible and applies an element-wise addition to fuse these two feature maps.
The dimentionality reduction is performed by 2D spatial average pooling, followed by 3D $1\times1\times1$ convolution. The details of this multi-scale aggregation operator is illustrated in Figure \ref{fig:3}.

\subsubsection{Spatial-attention propagation}
It has been shown that self attention from the later stage feature maps is an effective way to aggregate semantic context from across the scene and focus on the key elements of the scene \cite{wang2017residual}. We leverage this idea by using activation maps on the later stage feature maps to weight the early stage features. This guides our CIDC module to focus on the temporal evolution of the semantically important parts of the video clip.
We propose to propagate attention maps from later stage feature maps to early stage ones, and use them to generate task-attentive feature maps before they are fed to CIDC network. Formally,  
\begin{equation}
	\mathbf{F}'_{x}=\text{Bilinear}(\text{att}(\mathbf{F}))\odot \mathbf{F}_{x} + \mathbf{F}_{x}
\label{eqn:attn}
\end{equation}
\noindent where $\mathbf{F} \in \mathbb{R}^{C \times T \times W \times H}$ and $\mathbf{F}_{x}$ corresponds to later stage and early stage feature maps respectively, $\odot$ is the element-wise multiplication operator, $att(\mathbf{F}_{(t,i,j)} = sigmoid(mean(\mathbf{F}_{(t,i,j)})$ and $Bilinear$ denotes the bilinear interpolation operation that is used to upsample the spatial attention maps.

\subsection{Implementation Details}
\label{sec:impl_dets}
\subsubsection{Efficient CIDC Unit}
While conceptually our CIDC unit is straight forward, if implemented as separate convolutions for each time step (sub-clip feature) it would be very inefficient. Instead we use the intuition that our unit is equivalent to a standard $1 \times 1$ convolution with the upper triangle portion of the weights set to zero. However simply performing this operation poses issues for stable learning through SGD, and so here we present our normalization strategy to perform this operation efficiently, while keeping it stable during training. 

To be as efficient as possible, we would ideally construct the weights of convolution kernel ($\mathbf{w}$ from eq. \ref{eq:temp_ind_conv}) in such a way that $t$-th row of $\mathbf{w}_t^c$ abides by the conditions of eq. \ref{eq:temp_ind_conv}: $\mathbf{w}_t^c(t+1:T)~=0$ and that is compatible with SGD optimization. In other words, eq. \ref{eq:temp_ind_conv} is a differentiable operation that is compatible with the SGD family of training. 
To achieve this we take advantage of the fact that we are using a softmax operation to normalize each row $\mathbf{w}_t^c$. Modify eq. \ref{eq:temp_ind_conv} as follows, adding $-\inf$ to each element of $\mathbf{w}_t^c[t+1:T]$:
\begin{align}
	\mathbf{F}_c' &=  \left[\mathbf{w}_t^c * \mathbf{F}_c \right]_{t\in{1..T}} \\
	\mathbf{w}_t^c &=\text{softmax}(\mathbf{k}_t^c - [\inf \cdot \ \text{triu}(\mathbf{k})]_t^c)
\end{align}

\noindent where $\mathbf{k}$ is the learnable parameters, $\mathbf{k}_t^c$ is the $t^\text{th}$ row of $\mathbf{k}^c$, and $\mathbf{w}$ is computed from $\mathbf{k}$ that are passed through a softmax operation.
% , which ensures that the weights gets normalzied before given to each time point $t$ in $\mathbf{F}_c'$. 
By adding $-\inf$ to the upper triangle matrix of $\mathbf{k}$ we mask the convolution kernel and achieve our directional temporal convolution. 
We further linearly normalize the numerical values of weights in the lower-triangular portion of kernel $\mathbf{w}$ to the range of [-1, 1].
When $T' \neq T$, we first generate a square upper triangle matrix and rescale it to $T' \times T$ using bilinear interpolation. Finally, the uni-directional convolution can be extended to bi-directional by flipping the input data along temporal dimension as is done in bi-directional LSTMs \cite{schuster1997bidirectional}.

\subsubsection{Training Details}
% We first fine-tune the backbone network, and then train our multi-scale CIDC network by freezing the backbone.
We first pre-train the backbone network or initialize backbone from pre-trained models, and then train our multi-scale CIDC network.
We start training with a learning rate of 0.01, and decay it by a factor of 10 at epoch 40 and 80 respectively. In total the model is trained for 100 epoches.
We use stochastic gradient descent (SGD) with momentum of 0.9 and weight decay of $1e-4$ to train the network. 
In order to mitigate over-fitting, we add a dropout layer of rate of 0.6 for fully connected layers in CIDC network.
During training, we sample a random clip of length of 32 frames by skipping ever other frame (on average 15 fps).
We perform scale augmentation by randomly resizing the shorter size of training clip to between 256 and 320 pixels, and then randomly crop a video clip with spatial size of 224 $\times$ 224 pixels. Meanwhile, we apply random horizontal flips to the video clip at 0.5 probability.

During inference, we follow \cite{feichtenhofer2018slowfast} to first uniformly sample 10 clips from each video, and then resize the shorter side of every testing clip to 256 pixels. We then uniformly take three crops (with spatial size of 256 $\times$ 256) along its longer side, and finally we derive the prediction by taking average of predictions of all these 30 clips. 
Our experiments are conducted using the pytorch framework.

\section{Experiments}
\label{sec:exp}
\subsection{Dataset}
We evaluate our model on four commonly used datasets. 

\noindent \textbf{UCF 101 \cite{soomro2012dataset}} includes 101 categories of human actions. It contains more than 13K videos with an average length of 180 frames per video. 
Following previous works \cite{lin2018temporal,wang2016temporal}, we report the top-1 classification accuracy on the validation videos based on split 1. 

\noindent \textbf{HMDB 51 \cite{kuehne2011hmdb}} has a total of 6766 videos organized as 51 distinct action categories. 
The dataset has three splits and we report the top-1 classification accuracy on split 1 by following previous works \cite{lin2018temporal,wang2016temporal}.

\noindent \textbf{Something something V2 \cite{goyal2017something}} dataset consists of 174 actions and contains approximately 220,847 videos. Following other works \cite{wang2016temporal}, we report top-1 and top-5 classification accuracy on validation set. Something-Something dataset requires strong temporal modeling because many activities cannot be inferred based on spatial features only (e.g. open something, Covering something with something). 

\noindent \textbf{Kinetics 400 \cite{kay2017kinetics}} consists of approximately 240k training and 20k validation videos videos trimmed to 10 seconds from 400 human action categories.  Similar to other works, we report top-1 and top-5 classification accuracy on validation set.

\begin{table}
	\small
	\caption{Result comparison on UCF101 and HMDB51 datasets. We only compare with methods that use ResNet-50 as backbone and take as input the RGB video. All models are pre-trained on Kinetics-400 dataset.}
	\begin{center}
		\begin{tabularx}{0.65\textwidth}{l|ccccc} 
		    \toprule
			Model                 & Conv   &FLOPs  &Param. & HMDB51   & UCF101   \\ 
			\midrule
			R2D \cite{he2016deep}                     & 2D   &42G &24M            & 69.0    & 92.6 \\ 
			R2D-NL \cite{wang2018videos}              & 2D   &64G &31M            &72.5     &93.3 \\ 
			TSN \cite{wang2016temporal}               & 2D   &19G &11M            & 64.7    & 91.7 \\ 
			TSM \cite{lin2018temporal}                & 2D   &64G &24M            & 73.5    & 95.9 \\ 
			\hdashline
			I3D    \cite{wang2018videos}              & 3D   &65G &44M            & 69.1    & 92.9 \\ 
			I3D-NL \cite{wang2018videos}              & 3D   &94G &62M            & 72.2    &94.6 \\ 
			\midrule
			Ours (R2D)                                & 2D   &72G &85M            & 72.6    & 95.6 \\ 
			Ours (R2D-NL)                             & 2D   &91G &90M            & 73.3    & 95.9 \\
			\hdashline
			Ours (I3D)                                & 3D   &92G &87M            & \textbf{74.9}  & \textbf{97.2} \\ 
			Ours (I3D-NL)                             & 3D   &121G &103M          &  \textbf{75.2} & \textbf{97.9} \\ 
			\bottomrule
		\end{tabularx}
	\end{center}
	\label{tab:sota_ucf_hmdb}
\end{table}

\subsection{Comparison with state-of-the-art} 

\paragraph{UCF101 and HMDB51} 
We summarize the results in Table \ref{tab:sota_ucf_hmdb}. Our proposed method achieves state-of-the-art performance on both datasets. Our method improves upon both 2D and 3D baselines, both with and without non-local attention blocks. We see the biggest error reduction using our method on 3D networks, where our method reduces error by 11\% to 19\% on HMDB51 and {\bf 61\%} on UCF101.

\paragraph{Kinetics-400} 
We compare our model with state-of-the-art methods in Table \ref{tab:sota_kinetics}. %As our method is based on ResNet-50, we only compare with those that uses the same backbone network.  
The results for non-local \cite{wang2018non} and Slowfast network \cite{feichtenhofer2018slowfast} are obtained by running the model definition and weights provided by Gluon CV on our copy of the Kinetics 400 dataset. It is important to note that there is a consistent performance discrepancy between our reproduced results and those reported in \cite{wang2018non,feichtenhofer2018slowfast}. We believe that this is due to inconsistencies in the data as videos go missing from Kinetics 400 over time. 
The results in Table \ref{tab:sota_kinetics} show that the proposed multi-scale CIDC network again, consistently improves upon state-of-the-art methods for both 2D and 3D networks.

\begin{table}
\parbox{.45\textwidth}{
\caption{{ Result comparison on Kinetics-400 dataset. For fair comparison, we only compare with methods that use ResNet-50.}}
\centering
\begin{tabularx}{0.5\textwidth}{lcccc} 
		    \toprule
			Model        & Conv & FLOPs & Top1 & Top5  \\ 
			\midrule
			TSN \cite{wang2016temporal}    & 2D   &19G       & 70.6 & 89.2 \\ 
			R2D \cite{wang2018non}         & 2D   &42G       & 70.2 & 88.7 \\
			R2D-NL \cite{wang2018non}      & 2D   &64G       & 72.4 & 89.8 \\
			\hdashline
			I3D\cite{wang2018non}          & 3D   &65G       & 73.8 & 1.1 \\
			I3D-NL \cite{wang2018non}      & 3D   &94G       & 75.2 & 91.9 \\
			TSM \cite{lin2018temporal}     & 2D   &65G       & 74.1 & 92.2  \\ 
			bLVNet \cite{fan2019blvnet}    & 3D   &93G       & 74.3 &91.2 \\
			SF $4\times 16$ \cite{feichtenhofer2018slowfast}             & 3D   &36G        & 75.3   & 91.1  \\ 
			\midrule
			Ours (R2D)                     & 2D   &72G        & 72.2 & 90.1 \\
			Ours (R2D-NL)                  & 2D   &91G        & 72.8 & 90.5  \\
			\hdashline
			Ours (I3D)                     & 3D   &92G        & 74.5 & 91.3  \\
			Ours (I3D-NL)                  & 3D   &121G       & 75.6 & 92.4  \\
			Ours (Slowonly)                & 3D   &101G        & 75.5 & 92.1          \\
			\bottomrule
		\end{tabularx}
		\label{tab:sota_kinetics}
}
\hfill
\parbox{.45\textwidth}{
\caption{Result comparison on Something-Something V2 dataset. We only compare with methods that use ResNet-50 as backbone and take as input the RGB video. MS-TRN stands for multi-scale TRN and TS-TRN denotes two-stream TRN.}
\centering
\begin{tabularx}{0.47\textwidth}{lcccc} 
		   \toprule
			Model                 & Conv   & FLOPs   & Top1   & Top5   \\ 
			\midrule
			TSN \cite{wang2016temporal,lin2018temporal}   & 2D &19G  & 30.0   & 60.5  \\ 
			MS-TRN \cite{zhou2018temporal}                & 2D &33G  & 48.8   & 77.6  \\ 
			TS-TRN \cite{zhou2018temporal}                & 2D &42G  & 55.5   & 83.1  \\ 
			Fine-grain \cite{martinez2019action}          & 3D &69G  & 53.4   & 81.1 \\
			TSM \cite{lin2018temporal}                    & 2D &65G  & 63.4   & 88.5    \\ 
			bLVNet \cite{fan2019blvnet}                   & 3D &48G  & 61.7   & 88.1 \\
			R2D \cite{wang2018non}                        & 2D &42G  & 35.5   & 65.4  \\ 
			I3D\cite{wang2018non}                         & 3D &65G  & 49.6   & 78.2        \\ 
			\midrule
			Ours (R2D)                                    & 2D &72G  & 40.2   & 68.6         \\ 
			Ours (I3D) 		                              & 3D &92G  & 56.3   & 83.7     \\ 
			\bottomrule
		\end{tabularx}
% 	\end{center}
	\label{tab:sota_something_something}
}
\end{table}

\paragraph{Something-something V2} 
We compare our method with state-of-the-art methods in Table \ref{tab:sota_something_something}. Our method achieves very competitive results. By digging into the results, we observe that the proposed multi-scale CIDC network reduces the error on the baseline networks (R2D, I3D) by {\bf 7.3\%} and {\bf 13.2\%} respectively. This demonstrates that CIDC network learns important temporal information that is complementary to 3D convolutions.

\begin{table}[t]
\caption{Ablation experiments on UCF-101, HMDB-51 and Kinetics-400 (K400) datasets. Top-1 classification accuracy is reported.}
    \small
	\centering
	\subfloat[Result comparison with CIDC networks with different configurations.]{
		\begin{tabularx}{0.53\textwidth}{lccc} 
		    \toprule
			Model        & UCF   & HMDB & K400\\ 
			\midrule
			I3D-50                & 92.9  & 69.1 & 73.8     \\ 
			\midrule
			+ single-scale CIDC              & 95.2  & 73.6 & 74.0  \\ 
			+ Multi-scale CIDC               & 95.9  & 74.1 & 74.4  \\ 
			+ Spatial attention              & 97.2  & 74.9 & 74.5 \\ 
			\bottomrule
		\end{tabularx}
	}\hspace{2pt}
% 	\subfloat[Result comparison of multi-scale CIDC networks with different directional modeling units. ]{
	\subfloat[Result comparison of using different directional modeling units. ]{
		\begin{tabularx}{0.42\textwidth}{lccc} 
		   \toprule
			Model        & UCF & HMDB & K400   \\ 
			\midrule
			I3D-50                    & 92.9  & 69.1 & 73.8     \\ 
			\midrule
			non-direction             & 94.1  & 72.5 & 73.9  \\ 
			uni-direction             & 95.5  & 73.1 & 74.2 \\ 
			bi-direction              & 97.2  & 74.9 & 74.5 \\ 
			\bottomrule
		\end{tabularx}
	}
	\label{tab:ablation_1}
\end{table}

\subsection{Ablation Study}
We carefully perform the ablation study on one of the most challenging and largest-scale action classification dataset -- Kinetics-400 as well as on UCF-101 and HMDB-51 datasets. To facilitate the studies, we adopt I3D-50 as our feature backbone 
%\bing{I'm not sure whether it's a good idea or not to adopt I3D as backbone, as the performance gain is limited in this situation.}, 
unless specified. Results are summarized in Table \ref{tab:ablation_1}.

\paragraph{CIDC multi-scale is effective.}
We first look at the effect that the multi-scale and spatial attention have on the model performance.
We attach a single-scale CIDC network on top of the final feature map produced by the I3D-50 backbone; then we add our multi-scale version; and finally add our spatial attention.
We present these results in Table \ref{tab:ablation_1}(a). Each component provides a non-trivial boost in the final performance. These results show that by substituting single CIDC with its multi-scale alternative (w/ or w/o spatial attention propagation), we observe a healthy performance boost, which demonstrates the benefit of aggregating early stage feature maps.

\paragraph{Directional temporal modeling is important.} 
In order to understand the significance of directional temporal modeling, we instantiate the directional convolution in CIDC unit with three different configurations: (1), non-direction, where there is no temporal masking applied to the temporal convolution kernel. (2), uni-direction, where we apply temporal masking to the convolution kernel to make temporal modeling directional. and (3), bi-direction, where we apply the directional temporal modeling to both feature and temporally inverted feature, and concatenate the feature together along temporal axis. The performance of their corresponding multi-scale CIDC networks are listed in Table \ref{tab:ablation_1}(b). As shown, the model with bi-directional CIDC unit performs best, and it reduces error relative to its non-directional alternative by a significant percentage: 52.5\% 8.7\% 2.2\% on UCF-101, HMDB-51 nad Kinetics-400 respectively. These results validate the importance of directional temporal modeling in activity recognition. 
We notice that the performance improvement on UCF and HMDB dataset is more significant than that on Kinetics. We conjecture that it is because many videos in kinetics can be simply recognized by spotting key objects, thus undercutting the benefits of directional temporal modelling. 
% \bing{Someone double check this argument.}

\begin{table}[t]
    \caption{Ablation experiments on UCF-101, HMDB-51 and Kinetics-400 datasets. Top-1 classification accuracy is reported.}
    \small
	\centering
	\subfloat[Result comparison among different temporal modeling methods.]{
		\begin{tabularx}{0.45\textwidth}{lrrr} 
			\toprule
			Model        & UCF & HMDB  & Kinetics \\ 
			\midrule
			I3D                & 92.9 & 69.1 & 73.8      \\ 
			\midrule
			LSTM                 &  63.2  &31.3 & 63.4  \\ 
			self-attention       & 94.7  & 69.7 & 74.2  \\ 
			CIDC                 & 95.3  & 73.6 & 74.5  \\ 
			\bottomrule
		\end{tabularx}
	}  \hspace{5pt}
	\subfloat[Result comparison among feature fusion functions $\mathcal{S}$ on Kinetics-400.]{
		\begin{tabularx}{0.35\textwidth}{ll} 
			\toprule
			Model        & Acc\%   \\ 
			\midrule
			I3D                & 73.8        \\ 
			\midrule
			concatenate along $t$                 &  74.5      \\ 
			concatenate along $c$               & 74.1        \\ 
			sum                         & 73.7 \\
			\bottomrule
		\end{tabularx}
	}
	\label{tab:ablation_2}
\end{table}

\paragraph{Other temporal modeling methods are less effective.} 
We compare against two related clip-level temporal modelling methods -- self-attention and LSTM in Table \ref{tab:ablation_2}. In detail, we attach a network with 2-layer LSTM for temporal features with 512 LSTM unit on each layer (by performing 2D spatial pooling layer on spatial-temporal feature map $\mathbf{F}$ in Equation \ref{eq:temp_ind_conv}). Meanwhile, we attach a network with 2-layer self-attention block on top of spatial-temporal feature map $\mathbf{F}$. We used the vanilla self-attention \cite{vaswani2017attention}. Following previous work \cite{wang2018non}, we first flatten the spatio-temporal feature map and then use 3D convolution instead of linear layer used in \cite{vaswani2017attention} for linear projection.  
Their outputs are concatenated accordingly with the spatial-temporal feature map $\mathbf{F}$ before and after global pooling.  As shown in Table \ref{tab:ablation_2} (a), LSTM network does not perform well, even being outperformed by baseline I3D. This result is consistent with the observations in Carreira etc. \cite{carreira2017quo}. 
Even though self-attention network improves over I3D, it trails behind our proposed CIDC network. These results demonstrate that the proposed network is effective at learning clip-level motion information for action recognition. 

\paragraph{Feature fusion function $\mathcal{S}$} We experiment with different feature fusion functions $\mathcal{S}$ in Equation \ref{eqn:fusion}, and we summarize their results on Kinetics-400 in Table \ref{tab:ablation_2} (b). Overall, feature concatenation across temporal dimension $F_{out} = [F; F_{CIDC}]_t$ performs the best. We thus use this feature fusion function in our CIDC network.

\paragraph{Performance across different backbones.}
As shown in Table \ref{tab:sota_ucf_hmdb}, \ref{tab:sota_kinetics}, and \ref{tab:sota_something_something}, the proposed multi-scale CIDC network substantially boosts the top-1 accuracy both for 2-D (R-2D, R2D-NL) and 3-D (R-3D, I3D-NL, Slowfast) feature backbones on all four datasets. It shows that the temporal associations learned by the proposed CIDC network generalizes well to different state-of-the-art activity recognition models. It is particularly important to note that our method improves upon even backbones that contain similar temporal mechanisms (R2D-NL, I3D-NL). This shows that not only is our temporal modeling strategy powerful but also complementary to other common temporal modeling techniques.

\begin{table}[t]
\caption{ Quantitative analysis on Kinetics-400 dataset. The performance gain is defined as the disparity of the top-1 accuracy between CIDC network and that of I3D.}
    \footnotesize
	\centering
	\subfloat[Top 5 activity classes that are positively and negatively impacted by introducing CIDC network over I3D.]{
		\begin{tabularx}{0.8\textwidth}{llll} 
			\toprule
			Top 5 (+)        & Accuracy gain & Top 5 (-) & Accuracy gain   \\ 
			\midrule
			waxing legs  & +24\%  & kissing & -17\%     \\ 
			celebrating     &   +23\%  &garbage collecting &    -17\%  \\ 
			rock scissors paper &  +22\% & strumming guitar  & -17\%   \\ 
			climbing tree    &   +21\%  & yawning &   -16\%  \\ 
			ironing & +20\% & springboard diving & -14\%\\
			\bottomrule
		\end{tabularx}
	} \hfill
	\subfloat[The activity recognition accuracy gains of attaching CIDC network to I3D for activity pairs which share similar visual appearance.]{
		\begin{tabularx}{0.8\textwidth}{ll} 
			\toprule
			Activity pair        & Accuracy gain   \\ 
			\midrule
            waxing legs / shaving legs                 &  +24\% / +11\%      \\ 
			(swimming) breast stroke / butterfly stroke                 &  +22\% / +9\%      \\ 
			washing hair / curling hair                  &  +20\% / +11\%      \\ 
			long jump / triple jump                 &  +10\% / +6\%      \\ 
			bending metal / welding                 & +12\% / +8\% \\
			\bottomrule
		\end{tabularx}
	}
	\label{tab:error_anal_1}
\end{table}

\subsection{Error Analysis}
% \begin{figure}[t]
% 	\begin{center}
% 		\includegraphics[width=1\linewidth]{fig/fig6.png}
% 	\end{center}
% 	\caption{
% 	We show two video clips for action ``\texttt{waxing legs}" and ``\texttt{butterfly stroke}" respectively. Video examples are from Kinetics-400 dataset.}
% 	\label{fig:4}
% \end{figure}

\begin{figure*}[t]
	\begin{center}
		\includegraphics[width=0.95\linewidth]{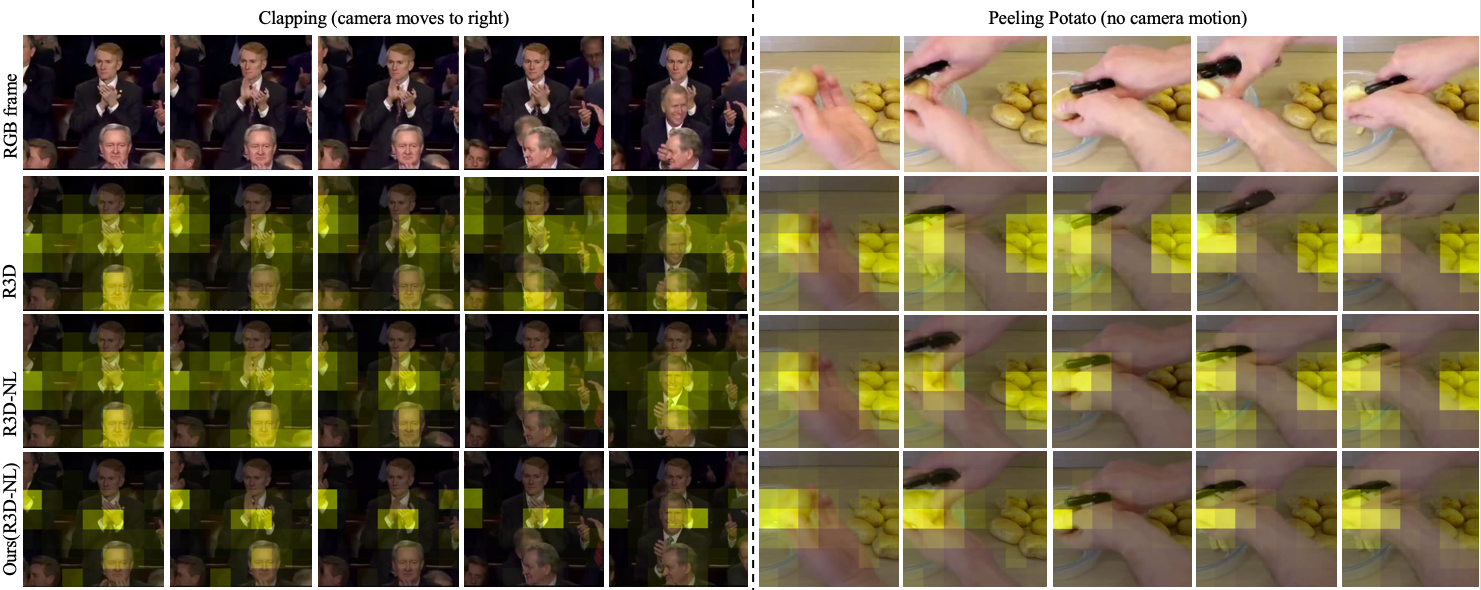}
	\end{center}
	\caption{
	We show two video clips (32 frames) and the spatial activation maps for the representative frame of every sub-clip (8 frames). I3D and I3D-NL network not only activates related image regions of ``\texttt{clapping}" and ``\texttt{peeling}", but also activates image regions that include irrelevant ``\texttt{people moving}" due to camera motion and irrelevant potato in the background respectively. In contrast, our proposed model  largely focuses on image region of interest that explains the action ``\texttt{clapping}" and ``\texttt{peeling}". Examples are from Kinetics-400 dataset.}
	\label{fig:5}
\end{figure*}
In order to understand which classes are impacted most by the proposed method, we compare per-class errors on Kinetics 400 dataset between I3D and our model.
In Table \ref{tab:error_anal_1} (a), we show the 5 action classes that are most positively and negatively impacted. We observe that our model improves the recognition performance for actions that exhibit large target object motions, e.g. ``\texttt{waxing legs, climbing trees}, whereas the model is confused for those actions that involves less obvious target object motions, e.g. ``\texttt{strumming guitar, yawning}". Given that the proposed model is to learn the clip-level temporal information, it's easier for the model to differentiate actions that exhibit large motions. We noticed the ``\texttt{garbage collection}" and ``\texttt{springboard diving}" should have noticeable motion but didn't benefit from our CIDC module. After watch the videos in these classes, we noticed the ``\texttt{garbage collection}" is often related to the garbage truck rather than the collection motion and the diving often has the camera motion with the athlete which makes the motion subtle. 
We also explore how our model performs for several challenging activity pairs whose visual appearance looks similar except the motion patterns exhibited by the targets are easily distinguishable (e.g. ``\texttt{swimming breast stroke}" vs ``\texttt{swimming butterfly stroke}"). As results shown in Table \ref{tab:error_anal_1} (b), our model significantly improves the recognition accuracy for all those activities over I3D. 
% We also show two visual examples in Figure \ref{fig:4} that were mis-predicted by I3D but correctly predicted by our proposed method.

\subsection{Visualizing CIDC activations}
We visualize the feature maps from I3D, I3D-NL and our CIDC network on videos from Kinetics-400 dataset. 
We generate the spatial activation map based on $att(F)$ in Equation \ref{eqn:attn} and show some representative examples in Fig \ref{fig:5}. 
From the visualized spatial attention maps, we can infer that:
1. I3D is only able to attend to image regions that are related to understanding actions but does not pick out the specific action in the scene. The top right of Figure \ref{fig:5} illustrates this as image regions related to the object ``\texttt{potato}'' and the action ``\texttt{peeling}'' are both activated by I3D to detect the ``\texttt{peeling\_potato}" action, even though the potatoes that are highlighted have nothing to do with the pealing action.
2. I3D-NL further narrows down these attention maps to focus on image regions that highly correlate with the action labels. Take action ``\texttt{clapping}" in Figure \ref{fig:5} as an example, some of the background object ``\texttt{person}" that does not perform the action ``\texttt{clapping}" is deactivated.  
3. Finally as examples in Figure \ref{fig:5} show, the proposed CIDC network only activates image regions that can explain the actions (e.g. ``\texttt{clapping}" and ``\texttt{peeling}"). This demonstrates that the 3D convolution is sensitive to motion across adjacent frames, however, without clip level contextual information, the 3D convolution is not able to distinguish whether the motion is related to action. As a result, the 3D convolution is likely to pick up all of the moving target. The CIDC network learns clip-level motion and has the contextual information about action target and irrelevant motion and thus tend to focus better on the action related features.
4. On Figure \ref{fig:1} and Figure \ref{fig:5}, the activation maps show the spatial attention propagation fires on both background and action related regions, and thus it is the CIDC unit that is able to focus on action related regions only

\section{Conclusion}
\label{sec:discussion}

In this paper, we first introduce the channel independent directional convolution (CIDC) unit, which learns temporal association among local features in a temporally directional fashion. Thus, it is able to encode the temporal ordering information of actions into feature maps. Moreover, we propose a light-weight network (based on CIDC units) that models the video clip-level temporal association of local spatial/spatial-temporal features.  
We test our method on four datasets and achieved the state-of-the-art performance. 
Our ablation study validates that the proposed CIDC is more effective at temporal modelling in action recognition. Furthermore, we visualize the activation map of CIDC network and show that it generally focuses on moving target that performs the actions.

% \clearpage\mbox{}Page \thepage\ of the manuscript.

% This is the last page of the manuscript.
% \par\vfill\par
% Now we have reached the maximum size of the ECCV 2020 submission (excluding references).
% References should start immediately after the main text, but can continue on p.15 if needed.

% \clearpage\mbox{}Page \thepage\ of the manuscript.
% \clearpage\mbox{}Page \thepage\ of the manuscript.

% This is the last page of the manuscript.
% \par\vfill\par
% Now we have reached the maximum size of the ECCV 2020 submission (excluding references).
% References should start immediately after the main text, but can continue on p.15 if needed.

\clearpage
% ---- Bibliography ----
%
% BibTeX users should specify bibliography style 'splncs04'.
% References will then be sorted and formatted in the correct style.
%
\bibliographystyle{splncs04}
\bibliography{egbib}

\begin{thebibliography}{10}
\providecommand{\url}[1]{\texttt{#1}}
\providecommand{\urlprefix}{URL }
\providecommand{\doi}[1]{https://doi.org/#1}

\bibitem{bilen2016dynamic}
Bilen, H., Fernando, B., Gavves, E., Vedaldi, A., Gould, S.: Dynamic image
  networks for action recognition. In: Proceedings of the IEEE Conference on
  Computer Vision and Pattern Recognition. pp. 3034--3042 (2016)

\bibitem{kay2017kinetics}
{Carreira}, J., {Zisserman}, A.: Quo vadis, action recognition? a new model and
  the kinetics dataset. In: 2017 IEEE Conference on Computer Vision and Pattern
  Recognition (CVPR). pp. 4724--4733 (July 2017). \doi{10.1109/CVPR.2017.502}

\bibitem{carreira2017quo}
Carreira, J., Zisserman, A.: Quo vadis, action recognition. A new model and the
  kinetics dataset. CoRR, abs/1705.07750  \textbf{2}, ~3 (2017)

\bibitem{donahue2015long}
Donahue, J., Anne~Hendricks, L., Guadarrama, S., Rohrbach, M., Venugopalan, S.,
  Saenko, K., Darrell, T.: Long-term recurrent convolutional networks for
  visual recognition and description. In: Proceedings of the IEEE conference on
  computer vision and pattern recognition. pp. 2625--2634 (2015)

\bibitem{fan2019blvnet}
Fan, Q., Chen, C.F.R., Kuehne, H., Pistoia, M., Cox, D.: {More Is Less:
  Learning Efficient Video Representations by Temporal Aggregation Modules}.
  In: Advances in Neural Information Processing Systems 33 (2019)

\bibitem{feichtenhofer2018slowfast}
Feichtenhofer, C., Fan, H., Malik, J., He, K.: Slowfast networks for video
  recognition. In: Proceedings of the IEEE International Conference on Computer
  Vision. pp. 6202--6211 (2019)

\bibitem{feichtenhofer2016convolutional}
Feichtenhofer, C., Pinz, A., Zisserman, A.: Convolutional two-stream network
  fusion for video action recognition. In: Proceedings of the IEEE conference
  on computer vision and pattern recognition. pp. 1933--1941 (2016)

\bibitem{fernando2016rank}
Fernando, B., Gavves, E., Oramas, J., Ghodrati, A., Tuytelaars, T.: Rank
  pooling for action recognition. IEEE transactions on pattern analysis and
  machine intelligence  \textbf{39}(4),  773--787 (2016)

\bibitem{fernando2015modeling}
Fernando, B., Gavves, E., Oramas, J.M., Ghodrati, A., Tuytelaars, T.: Modeling
  video evolution for action recognition. In: Proceedings of the IEEE
  Conference on Computer Vision and Pattern Recognition. pp. 5378--5387 (2015)

\bibitem{girdhar2019video}
Girdhar, R., Carreira, J., Doersch, C., Zisserman, A.: Video action transformer
  network. In: Proceedings of the IEEE Conference on Computer Vision and
  Pattern Recognition. pp. 244--253 (2019)

\bibitem{goyal2017something}
Goyal, R., Kahou, S.E., Michalski, V., Materzynska, J., Westphal, S., Kim, H.,
  Haenel, V., Fruend, I., Yianilos, P., Mueller-Freitag, M., et~al.: The"
  something something" video database for learning and evaluating visual common
  sense. In: ICCV. vol.~1, p.~3 (2017)

\bibitem{he2016deep}
He, K., Zhang, X., Ren, S., Sun, J.: Deep residual learning for image
  recognition. In: Proceedings of the IEEE conference on computer vision and
  pattern recognition. pp. 770--778 (2016)

\bibitem{hochreiter1997long}
Hochreiter, S., Schmidhuber, J.: Long short-term memory. Neural computation
  \textbf{9}(8),  1735--1780 (1997)

\bibitem{karpathy2014large}
Karpathy, A., Toderici, G., Shetty, S., Leung, T., Sukthankar, R., Fei-Fei, L.:
  Large-scale video classification with convolutional neural networks. In:
  Proceedings of the IEEE conference on Computer Vision and Pattern
  Recognition. pp. 1725--1732 (2014)

\bibitem{krizhevsky2012imagenet}
Krizhevsky, A., Sutskever, I., Hinton, G.E.: Imagenet classification with deep
  convolutional neural networks. In: Advances in neural information processing
  systems. pp. 1097--1105 (2012)

\bibitem{kuehne2011hmdb}
Kuehne, H., Jhuang, H., Garrote, E., Poggio, T., Serre, T.: Hmdb: a large video
  database for human motion recognition. In: 2011 International Conference on
  Computer Vision. pp. 2556--2563. IEEE (2011)

\bibitem{li2017progress}
Li, X., Zhang, Y., Zhang, J., Zhou, M., Chen, S., Gu, Y., Chen, Y., Marsic, I.,
  Farneth, R.A., Burd, R.S.: Progress estimation and phase detection for
  sequential processes. Proceedings of the ACM on interactive, mobile, wearable
  and ubiquitous technologies  \textbf{1}(3), ~73 (2017)

\bibitem{li2018videolstm}
Li, Z., Gavrilyuk, K., Gavves, E., Jain, M., Snoek, C.G.: Videolstm convolves,
  attends and flows for action recognition. Computer Vision and Image
  Understanding  \textbf{166},  41--50 (2018)

\bibitem{lin2018temporal}
Lin, J., Gan, C., Han, S.: Tsm: Temporal shift module for efficient video
  understanding. In: Proceedings of the IEEE International Conference on
  Computer Vision (2019)

\bibitem{luo2016understanding}
Luo, W., Li, Y., Urtasun, R., Zemel, R.: Understanding the effective receptive
  field in deep convolutional neural networks. In: Advances in neural
  information processing systems. pp. 4898--4906 (2016)

\bibitem{martinez2019action}
Martinez, B., Modolo, D., Xiong, Y., Tighe, J.: Action recognition with
  spatial-temporal discriminative filter banks. Proceedings of the IEEE
  International Conference on Computer Vision  (2019)

\bibitem{perez2013tv}
P{\'e}rez, J.S., Meinhardt-Llopis, E., Facciolo, G.: Tv-l1 optical flow
  estimation. Image Processing On Line  \textbf{2013},  137--150 (2013)

\bibitem{schuster1997bidirectional}
Schuster, M., Paliwal, K.K.: Bidirectional recurrent neural networks. IEEE
  Transactions on Signal Processing  \textbf{45}(11),  2673--2681 (1997)

\bibitem{shen2018disan}
Shen, T., Zhou, T., Long, G., Jiang, J., Pan, S., Zhang, C.: Disan: Directional
  self-attention network for rnn/cnn-free language understanding. In:
  Thirty-Second AAAI Conference on Artificial Intelligence (2018)

\bibitem{simonyan2014two}
Simonyan, K., Zisserman, A.: Two-stream convolutional networks for action
  recognition in videos. In: Advances in neural information processing systems.
  pp. 568--576 (2014)

\bibitem{soomro2012dataset}
Soomro, K., Zamir, A.R., Shah, M.: A dataset of 101 human action classes from
  videos in the wild. Center for Research in Computer Vision  (2012)

\bibitem{sun2019deep}
Sun, K., Xiao, B., Liu, D., Wang, J.: Deep high-resolution representation
  learning for human pose estimation. In: CVPR (2019)

\bibitem{sun2019high}
Sun, K., Zhao, Y., Jiang, B., Cheng, T., Xiao, B., Liu, D., Mu, Y., Wang, X.,
  Liu, W., Wang, J.: High-resolution representations for labeling pixels and
  regions. CoRR  \textbf{abs/1904.04514} (2019)

\bibitem{taylor2010convolutional}
Taylor, G.W., Fergus, R., LeCun, Y., Bregler, C.: Convolutional learning of
  spatio-temporal features. In: European conference on computer vision. pp.
  140--153. Springer (2010)

\bibitem{tran2015learning}
Tran, D., Bourdev, L., Fergus, R., Torresani, L., Paluri, M.: Learning
  spatiotemporal features with 3d convolutional networks. In: Proceedings of
  the IEEE international conference on computer vision. pp. 4489--4497 (2015)

\bibitem{tran2018closer}
Tran, D., Wang, H., Torresani, L., Ray, J., LeCun, Y., Paluri, M.: A closer
  look at spatiotemporal convolutions for action recognition. In: Proceedings
  of the IEEE conference on Computer Vision and Pattern Recognition. pp.
  6450--6459 (2018)

\bibitem{vaswani2017attention}
Vaswani, A., Shazeer, N., Parmar, N., Uszkoreit, J., Jones, L., Gomez, A.N.,
  Kaiser, {\L}., Polosukhin, I.: Attention is all you need. In: Advances in
  neural information processing systems. pp. 5998--6008 (2017)

\bibitem{wang2017residual}
Wang, F., Jiang, M., Qian, C., Yang, S., Li, C., Zhang, H., Wang, X., Tang, X.:
  Residual attention network for image classification. In: Proceedings of the
  IEEE Conference on Computer Vision and Pattern Recognition. pp. 3156--3164
  (2017)

\bibitem{wang2016temporal}
Wang, L., Xiong, Y., Wang, Z., Qiao, Y., Lin, D., Tang, X., Van~Gool, L.:
  Temporal segment networks: Towards good practices for deep action
  recognition. In: European conference on computer vision. pp. 20--36. Springer
  (2016)

\bibitem{wang2018non}
Wang, X., Girshick, R., Gupta, A., He, K.: Non-local neural networks. In:
  Proceedings of the IEEE Conference on Computer Vision and Pattern
  Recognition. pp. 7794--7803 (2018)

\bibitem{wang2018videos}
Wang, X., Gupta, A.: Videos as space-time region graphs. In: Proceedings of the
  European conference on computer vision (ECCV). pp. 399--417 (2018)

\bibitem{wu2019long}
Wu, C.Y., Feichtenhofer, C., Fan, H., He, K., Krahenbuhl, P., Girshick, R.:
  Long-term feature banks for detailed video understanding. In: Proceedings of
  the IEEE Conference on Computer Vision and Pattern Recognition. pp. 284--293
  (2019)

\bibitem{xie2017aggregated}
Xie, S., Girshick, R., Doll{\'a}r, P., Tu, Z., He, K.: Aggregated residual
  transformations for deep neural networks. In: Proceedings of the IEEE
  conference on computer vision and pattern recognition. pp. 1492--1500 (2017)

\bibitem{xie1712rethinking}
Xie, S., Sun, C., Huang, J., Tu, Z., Murphy, K.: Rethinking spatiotemporal
  feature learning: Speed-accuracy trade-offs in video classification. In:
  Proceedings of the European Conference on Computer Vision (ECCV). pp.
  305--321 (2018)

\bibitem{yue2015beyond}
Yue-Hei~Ng, J., Hausknecht, M., Vijayanarasimhan, S., Vinyals, O., Monga, R.,
  Toderici, G.: Beyond short snippets: Deep networks for video classification.
  In: Proceedings of the IEEE conference on computer vision and pattern
  recognition. pp. 4694--4702 (2015)

\bibitem{zhou2018temporal}
Zhou, B., Andonian, A., Oliva, A., Torralba, A.: Temporal relational reasoning
  in videos. In: Proceedings of the European Conference on Computer Vision
  (ECCV). pp. 803--818 (2018)

\end{thebibliography}
\end{document}